%% file: Neuro-AIF-robotics-iclr2021.tex
\documentclass{article} 
\usepackage{iclr2021_conference,times}
\iclrfinalcopy
\input{math_commands.tex}

\usepackage{hyperref}
\usepackage{url}
\usepackage[center]{subfigure}
\usepackage{multirow}
\usepackage{graphicx}

\renewcommand{\v}[1]{{\boldsymbol{\mathbf{#1}}}}
\newcommand\latent{\v{z}}
\newcommand\obs{\v{x}}

\newcommand\action{\v{a}}

\newcommand\g{g}
\newcommand\f{f}
\newcommand\xnoise{\v{r}}
\newcommand\znoise{\v{w}}
\newcommand\joints{\v{q}}

\title{\Large{Neuroscience-inspired perception-action in robotics}\\
\normalsize{applying active inference for state estimation, control and self-perception}}

\author{Pablo Lanillos\\
Donders Institute for Brain, Cognition and Behaviour\\
Department of Artificial Intelligence\\
Radboud University\\
Nijmegen, the Netherlands \\
\texttt{\{p.lanillos\}@donders.ru.nl} \\
\AND 
Marcel van Gerven\\
Donders Institute for Brain, Cognition and Behaviour\\
Department of Artificial Intelligence\\
Radboud University\\
Nijmegen, the Netherlands \\
}

\begin{document}

\maketitle

\begin{abstract}
Unlike robots, humans learn, adapt and perceive their bodies by interacting with the world. Discovering how the brain represents the body and generates actions is of major importance for robotics and artificial intelligence. Here we discuss how neuroscience findings open up opportunities to improve current estimation and control algorithms in robotics. In particular, how active inference, a mathematical formulation of how the brain resists a natural tendency to disorder, provides a unified recipe to potentially solve some of the major challenges in robotics, such as adaptation, robustness, flexibility, generalization and safe interaction. This paper summarizes some experiments and lessons learned from developing such a computational model on real embodied platforms, i.e., humanoid and industrial robots. Finally, we showcase the limitations and challenges that we are still facing to give robots human-like perception\footnote{Accepted at ICLR 2021 Brain2AI workshop. Video presentation: \url{https://youtu.be/oW40kUGxu2s}}.
\end{abstract}

\section{Introduction}
\label{sec:intro}

\begin{quote}
    She wakes up, looks at the mirror and reflects --\textit{is this me?}-- while opening the tap and leaving the water to pour out --\textit{did I do it?}
\end{quote} 

Answering these two simple questions is the tip of the iceberg of body perception and action in the brain~\citep{tsakiris2007agency,haselager2013did,haggard2017sense,hinz2018drifting}. Underneath, there are more than 500 million years of neural development that allow us to safely interact in a world full of uncertainties. Unveiling how the brain integrates different sources of information to perceive the body, and generates adaptive actions is of major importance for robotics and artificial intelligence~\citep{todorov2002optimal,oliver2021empirical}.

According to \citep{Helmholtz1867} perception is an unconscious mechanism that infers the state of the world. Under this revolutionary view, the brain may learn an internal generative model of the world and use it to reconstruct the perceived reality from partial sensory information. A perfect example is visual illusions, such as the Dallenbach illusion, where prior information disambiguates the meaning of an altered noisy image~\citep{kmd1951puzzle}. In other words, the brain works as a predictive machine \citep{clark2013whatever}. This prediction power has been proposed for several segregated brain regions, such as the visual cortex~\citep{rao1999predictive} or the cerebellum~\citep{doya1999computations}. One way to model a predictive machine is through Bayesian inference, where the agent’s beliefs about the state of the environment are based on its sensory evidence and prior beliefs. Thus, the internal model is updated through the senses~\citep{knill2004bayesian,friston2005theory}. Accordingly, we can answer the question `Where is my body?' by estimating our body in space using learned models that have been acquired during our lifetime~\citep{mori2010human} based on afferent multisensory input (e.g., visual, tactile and proprioceptive cues).

Analogously, the generation of adaptive behavior from the predictive brain perspective arises from the minimization of surprise~\citep{friston2010unified}, that is, the difference between predicted and observed sensations. According to this predictive processing hypothesis---see \cite{ciria2021predictive} for a recent review in cognitive robotics---, the brain maintains an internal model that predicts the agent’s sensations based on the causes of those sensations in the environment. If we condition sensation not only on the environmental causes but also on the agent’s actions then we can integrate perception and action. Adaptive behavior can then be viewed as an active inference (AIF) process in which the agent selects those actions that support the maximization of model evidence, or equivalently, the minimization of surprise~\citep{da2020active}. Consequently, we can answer the question `How should I move my body?' by computing those actions that make the world better fit the learned model. 
These are the foundations of the free energy principle (FEP, \cite{friston2010unified}), which postulates that the brain optimizes the variational free energy (VFE), which is an upper bound on model evidence. This approach allows us to infer and simultaneously adapt body posture to uncertain situations \citep{kirchhoff2018markov}; a goal that has only been partially solved in robotics.

\begin{figure}[hbtp!]
	\centering
    \includegraphics[width=0.8\textwidth]{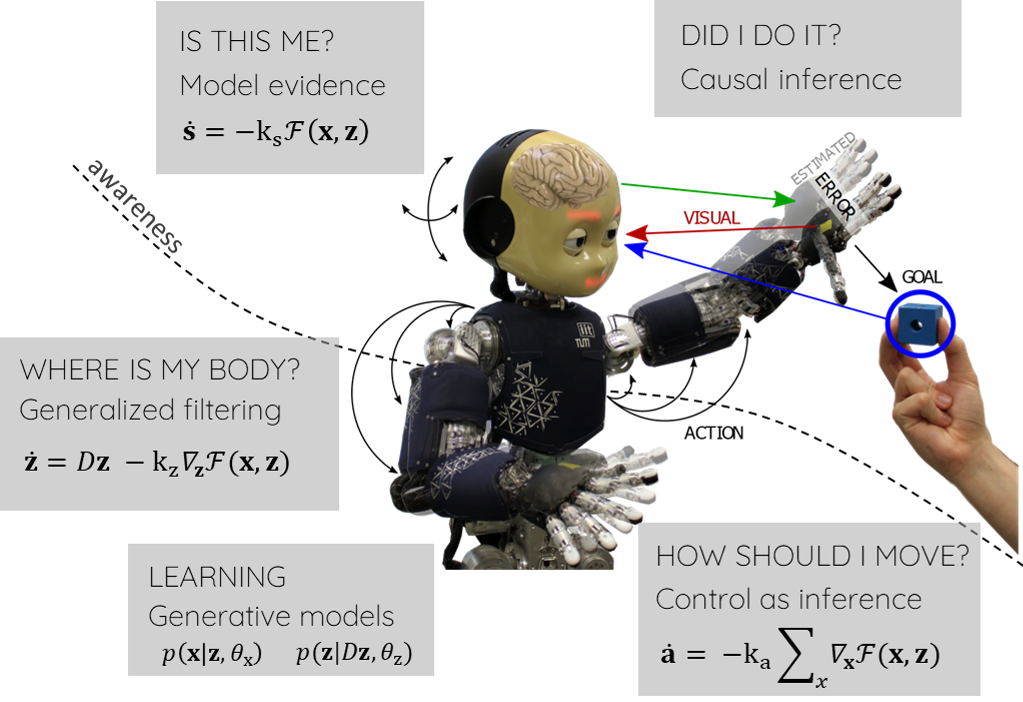}
	\caption{Neuroscience-inspired body perception and action for robotics. Answering these four questions is crucial for embodied artificial intelligence and robotics. We cast estimation (\textit{where is my body?}) and control (\textit{how should I move?}) as an inference process where the overall aim is to perform state estimation. While we already approached a basic form of body-ownership (\textit{is this me?}) directly from the model evidence, agency (\textit{did I do it?}) is still an evasive question.}
	\label{fig:questions}
\end{figure}

Recently, we have shown that combining active inference with advances in deep learning allows us to generate adaptive behavior in humanoid and industrial robots. The single aim of the robot is to infer its state (unobserved variable) by means of noisy sensory inputs (observed). For that purpose, it can refine its state using the measurements or perform actions to fit the observed world to its internal model. This is dually computed by optimizing the VFE. This work, depicted in Figure~\ref{fig:questions}, summarizes some of the recent proof-of-concept experiments where we studied coupled perception (estimation) and action (control) using deep AIF models~\citep{lanillos2018adaptive,sancaktar2020end,oliver2021empirical,lanillos2020predictive,meo2021multimodal}. Furthermore, the application of such a model in real-world settings provides an exciting avenue for mechanistically explaining neuropsychological observations related to behavior in biological agents.
\vspace{-0.09cm}

The following sections are organized to answer the questions described in Figure~\ref{fig:questions} from the active inference perspective.

\section{Where is my body? generalized filtering}
\label{sec:where}
Knowing where our body is in space is central for safety and awareness. The brain keeps track of a body posture or schema~\citep{hoffmann2010body, lanillos2017enactive} by fusing information from all available sensory inputs. We cast body state estimation as generalized filtering~\citep{friston2010generalised}. Defining the state of the system/body as $\latent$, the sensory inputs as $\obs$ and the time-derivative of the state vector $D\latent$ in generalized coordinates\footnote{ $D \latent = \frac{d}{dt}(z^{[0]},z^{[1]},\ldots,z^{[n]}) \!=\! [z^{[1]},z^{[2]},\ldots,z^{[n+1]}]$; where the superscript indicates the derivative order.}, we describe the generative model of the body/world with two functions:
\begin{align}
    \obs &= \g(\latent) + \xnoise  & \text{sensory input (visual, proprioceptive, etc)} \nonumber\\
    D\latent &= \f(\latent) + \znoise &\text{internal state dynamics} \nonumber
    \label{gen}
\end{align}

Body estimation is solved by inferring the system state with the following update equation:
\begin{equation}
   \dot{\latent} =  D\latent - k_{z}\nabla_{\latent}\mathcal{F}({\latent}, {\obs})
    \label{eq:perception_general}
\end{equation}
$\mathcal{F}$ is the VFE and under the Laplace and mean-field approximations \citep{friston2007variational,buckley2017free,oliver2021empirical}, has closed form and is defined as\footnote{Other approximations of the variational density are also possible but out of the scope of this paper~\citep{ambrogioni2021automatic}.}:
\begin{align}
    \mathcal{F}(\latent,\obs) \simeq& -\ln p(\latent,\obs) = -\ln \left(p(\obs|\latent) p(\latent)\right)\\
    \simeq& \;(\obs-g(\obs))^T \Sigma_x^{-1} (\obs-g(\obs)) \nonumber\\
    &+ (D\latent-f(\latent))^T \Sigma_z^{-1}(D\latent-f(\latent)) \nonumber\\
    &+ \frac{1}{2} \ln|\Sigma_x| + \frac{1}{2} \ln|\Sigma_z| \label{eq:flaplace}
\end{align}
Note that Equation (\ref{eq:perception_general}) is correcting the state given the weighted prediction error encoded in $\mathcal{F}$.

In~\citep{lanillos2018adaptive, oliver2021empirical}, we described how to use this method to estimate the body pose of a humanoid robot. We showed how the algorithm provided robust multisensory fusion (visual, proprioceptive and tactile sources) when injecting strong readings noise, and adaptability to unexpected sensory changes, such as visuo-tactile perturbations or broken sensors.

\subsection{Learning the generative models}
\begin{figure}[hbtp!]
	\centering
    \includegraphics[width=0.85\textwidth]{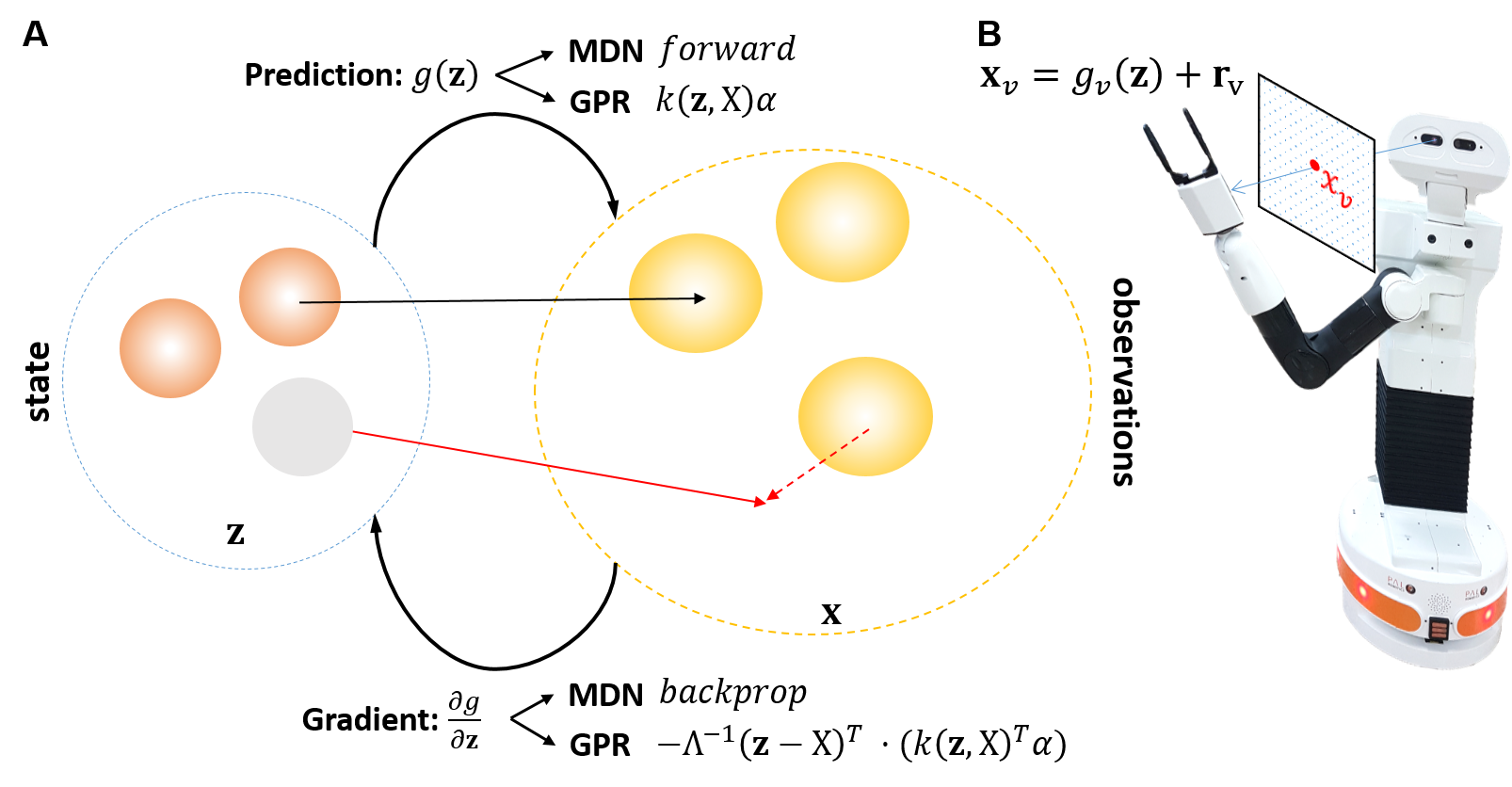}
	\caption{Sensory generative model learning. (A) Mapping between the latent state and the observations. Mixture density network (MDN) or Gaussian process regression (GPR) used to learn the kinematic mapping with low-dimensional inputs. While in GPR there is a close form for computing the prediction and the partial derivative with respect to the state, in a neural network approach we exploit the forward pass and the backpropagation algorithm. ($X$ refers to the training data, $k(\cdot)$ to the squared exponential kernel and $\alpha$ is a numerically stable computation of the inverse covariance). (B) Segmented end-effector position in the visual space and its associated generative model $g_v(\latent)$.}
	\label{fig:g_learning}
\end{figure}

Usually, within the AIF literature, the generative model of the body is known a priori~\citep{friston2010action}. However, in practice, the brain learns these models during interaction. We introduced several methods to learn the sensory generative model $g(\latent)$, i.e., the mapping between the state of the system and the observed measurements~\citep{lanillos2018active}, and to seamlessly include them in the VFE optimization scheme. Although these methods are not (yet) biologically plausible, we are currently mainly interested in the functional aspects of the model. For low-dimensional inputs, i.e., when we can segment the location of the end-effector in the image or in the task space, Gaussian process regression~\citep{rasmussen2005GPM} or mixture density networks~\citep{Bishop1994} are suitable. In the former, we can compute in closed form the partial derivatives of the generative functions with respect to the state \citep{lanillos2018active}. In the latter, we can exploit the backpropagation method to compute it~\citep{lanillos2020robot}. Figure \ref{fig:g_learning} describes how to learn the visual kinematic mapping that is needed to solve Eq.~(\ref{eq:perception_general}).

Learning the internal state dynamical generative model $f(\latent)$ is more complex \citep{lanillos2018active}. However, if we know the characteristics of the system we can use a simplified state model and let the variational approximation to tackle the discrepancies~\citep{friston2010action, pio2016active}. In the extreme case, if we have access to the propriceptive desired state, we can use a linear model~\citep{oliver2021empirical, meo2021multimodal}. The adaptive controller will absorb the non-linearities by tracking the desired joint positions and velocities~\cite{pezzato2020novel,meo2021multimodal}. 

\subsection{Scaling to high-dimensional inputs: pixel-AIF}

\begin{figure}[hbtp!]
	\centering
    \includegraphics[width=1.0\linewidth]{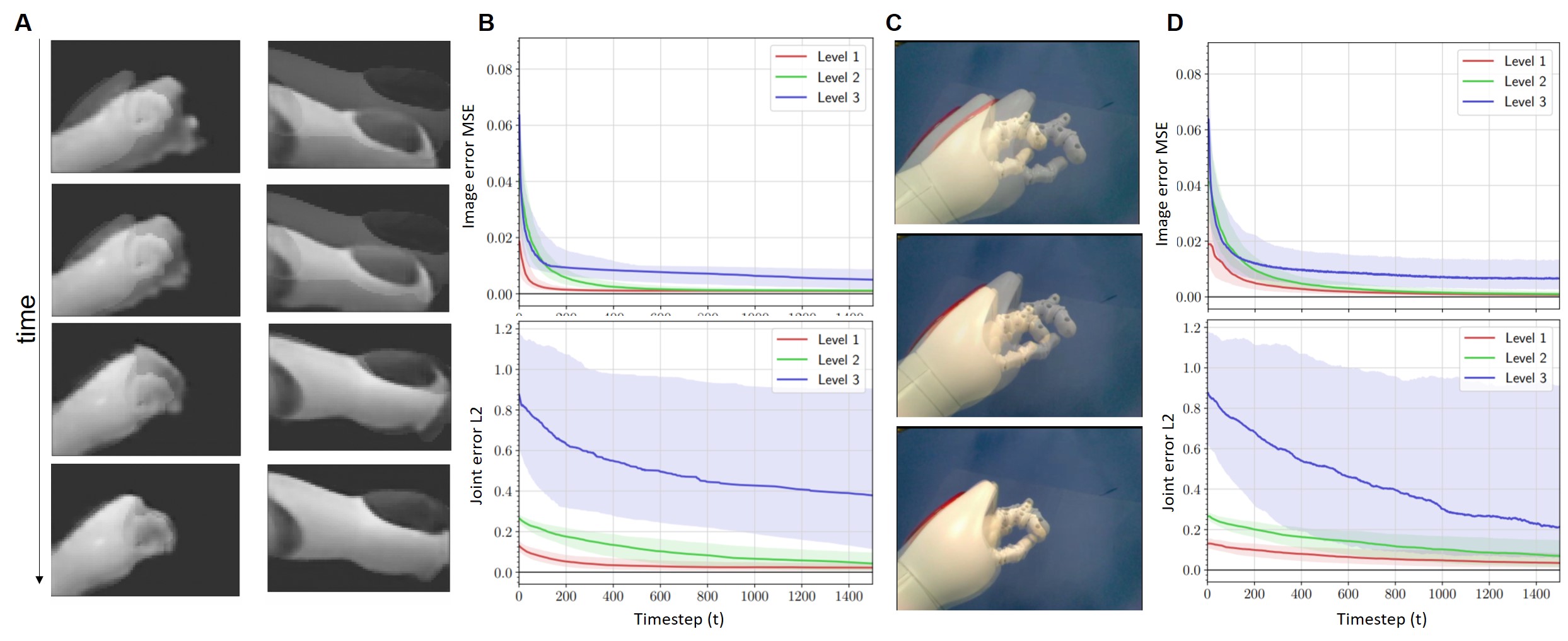}
	\caption{Pixel-AIF algorithm. Image reproduced from \citep{sancaktar2020end}. (A) Pixel-based body estimation. (B) Image and joint angles estimation average errors. We evaluated the arm state estimation for three different levels of difficulty: small deviations (red), strong deviations (green) and random pose (blue); computed over 2500 trials (i.e., 5$\times$ 500 random arm images). (C) Pixel-AIF working in the NAO robot using the head bottom camera in a reaching task. The desired goal is defined as an image with the arm in a random position. (D) Image and joint angles estimation average errors in the reaching task evaluation using the Pixel-AIF.}
	\label{fig:pixelAIF}
\end{figure}

By means of deep artificial neural networks, we can scale the state estimation to high-dimensional inputs. In~\citep{sancaktar2020end} we showed how to use a convolutional decoder to perform pixel state estimation. Figure \ref{fig:pixelAIF} shows the Pixel-AIF algorithm working for estimation and control in the NAO robot using only raw images as input. Figure \ref{fig:pixelAIF}A shows the evolution of the predicted arm image until the state converges to the right state estimation. Figure \ref{fig:pixelAIF}B shows the average errors of the state estimation from pixels in three different conditions: level 1, small deviations of the joint angles from the prior belief; level 2, strong deviations from the prior belief; and level 3 a random pose.

In \citep{meo2021multimodal}, we extended the Pixel-AIF framework to multimodal state representation learning. This implies the inclusion of one decoder per sensory input. One of the main advantages of our brain-inspired approach is that the robot only has to learn the kinematic forward mapping and then is the inference process that performs the Bayesian inversion for estimation and produces the right torques for achieving the goal, even in the presence of unmodeled situations or external forces.

\section{How should I move? Action as inference}
\label{sec:how}
In predictive brain models, such as AIF, action generation is realized in a unique manner. The brain infers actions in the light of anticipated sensory consequences. Thus, the system should encode desired goals (or preferences) as (learned) priors. These desired goal states (imaginary goals) can trigger corresponding, expected sensations which, in turn, initiate and control bodily movements through the reflex arc pathway. This is in line with ideomotor theory, which poses that to generate a motor response we first create a representation of the goal~\citep{James1890}.
We cast action as a control as inference problem where the action is updated through gradient-based optimization:
\begin{equation}
    \dot{\action} = -k_a \sum_x \frac{d \obs}{da} \cdot \nabla_{\obs}\mathcal{F}({\obs}, {\latent})
    \label{eq:action_general}
\end{equation}

\subsection{Emergence of the task by operational specification}
The advantage of the AIF approach is that the task can be specified as a desired preference. This implies that the prior will lead to the corresponding actions. The difficulty becomes finding the right attractor set. For instance, we can set the goal as the sensory output that we would like to obtain within the internal dynamics as follows:
\begin{align}
f(\latent,\obs_d) = T(\latent) (\obs_d -g(\latent)) = \frac{\partial \g(\latent)}{\partial \latent} (\obs_d - g(\latent))
\end{align}
where $T(\latent)$ is a function that maps the error in the sensory space to the latent space and the operational specification defines, for instance, the desired outcome image and joint angles $\obs_d=\{\v{I}_d, \joints_d\}$.

Figure \ref{fig:pixelAIF}C shows the NAO robot generating actions using the pixel-AIF~\citep{sancaktar2020end} until reaching the desired goal. Here the operational specification is only the final image. Figure \ref{fig:pixelAIF}D describes the statistical evaluation of the image and joint angle errors when enabling AIF control. While deviations (small or large) from the prior belief have good performance, some of the random images (level 3) were not fully solved---See the joint errors large variance of Fig.~\ref{fig:pixelAIF}D. This points out the intrinsic nature of this brain-inspired method. It is supposed to work for adaptive corrections of the body in the space but not for complex sequential goal-driven tasks. To solve planning the attractor set should be encoded as a prior or learned~\citep{tani2003learning}. A promising new research direction is to use an explicit policy and compute the expected free energy~\citep{millidge2021whence} approximating the density functions using artificial neural networks~\citep{van2020deep}.

\subsection{Adaptation: involuntary actions to fit a new world}


\begin{figure}[hbtp!]
\begin{minipage}{0.65\textwidth}
\centering
	\subfigure[Visual marker to the left.]{
		\centering
		\includegraphics[width=0.3\textwidth]{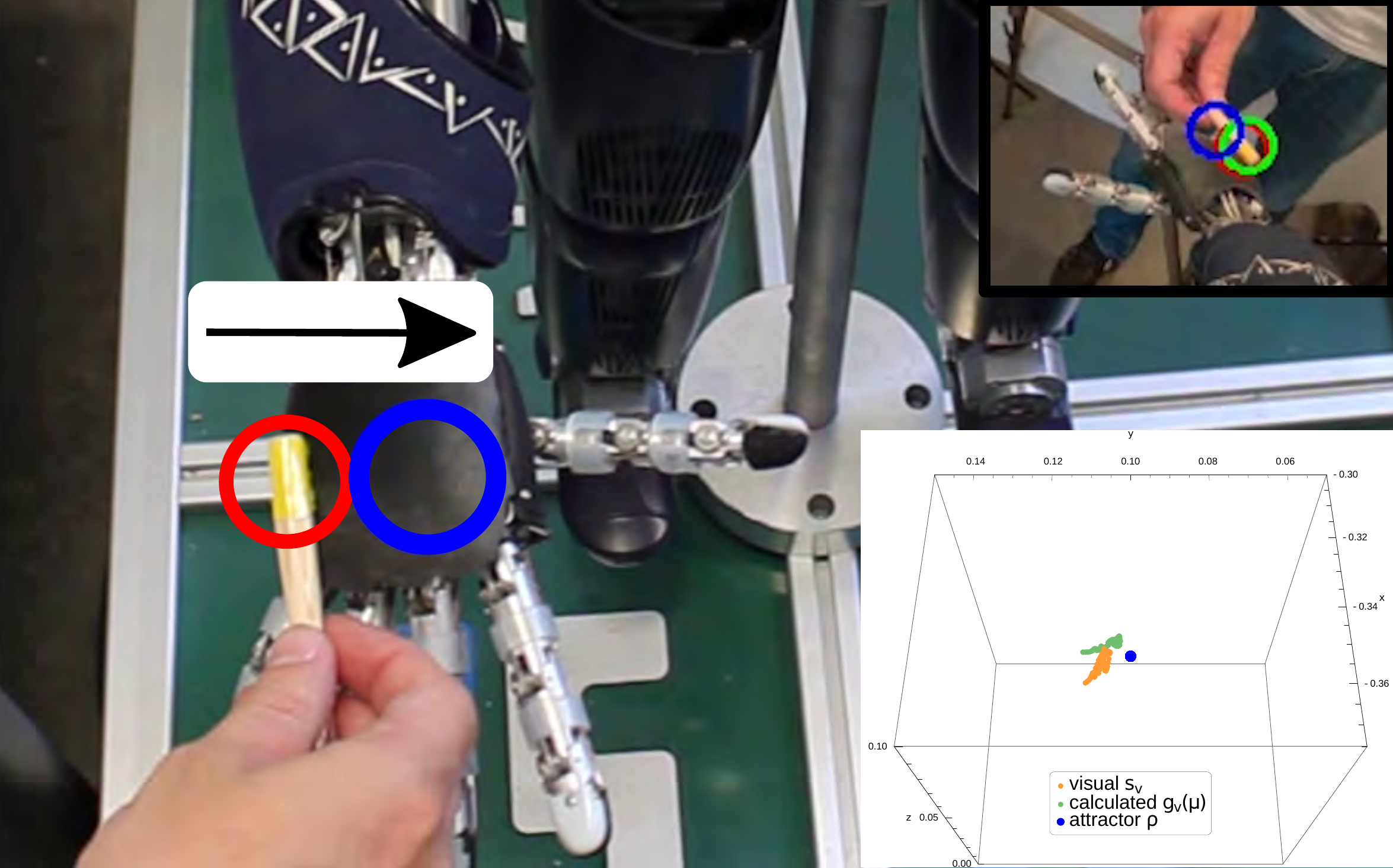}
		\label{results:adapt:left-deviation}
	}	
	\subfigure[Visual marker to the right.]{
		\centering
		\includegraphics[width=0.3\textwidth]{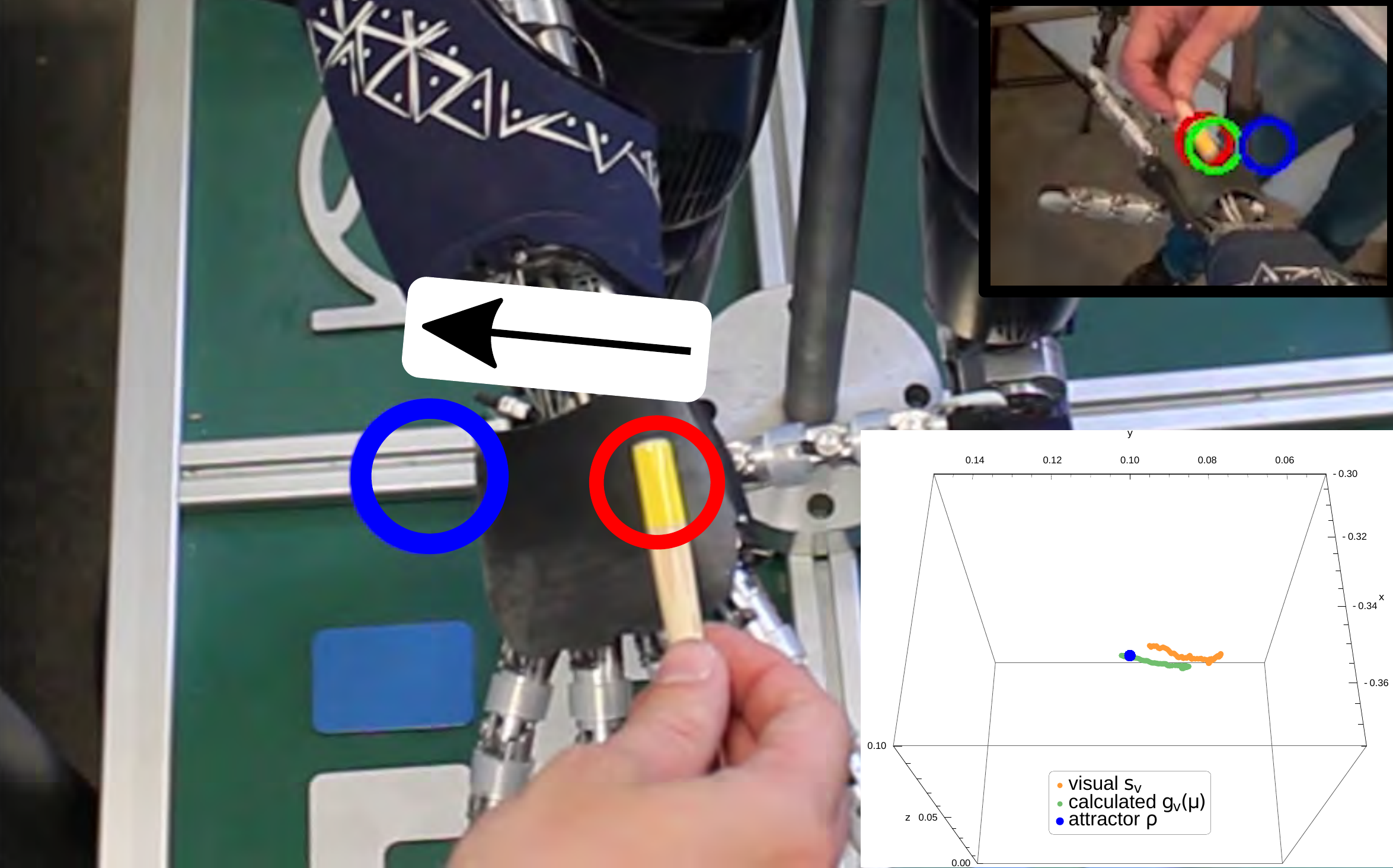}
		\label{results:adapt:right-deviation}
	}\\
	\subfigure[Visual marker on finger.]{
		\centering
		\includegraphics[width=0.3\textwidth]{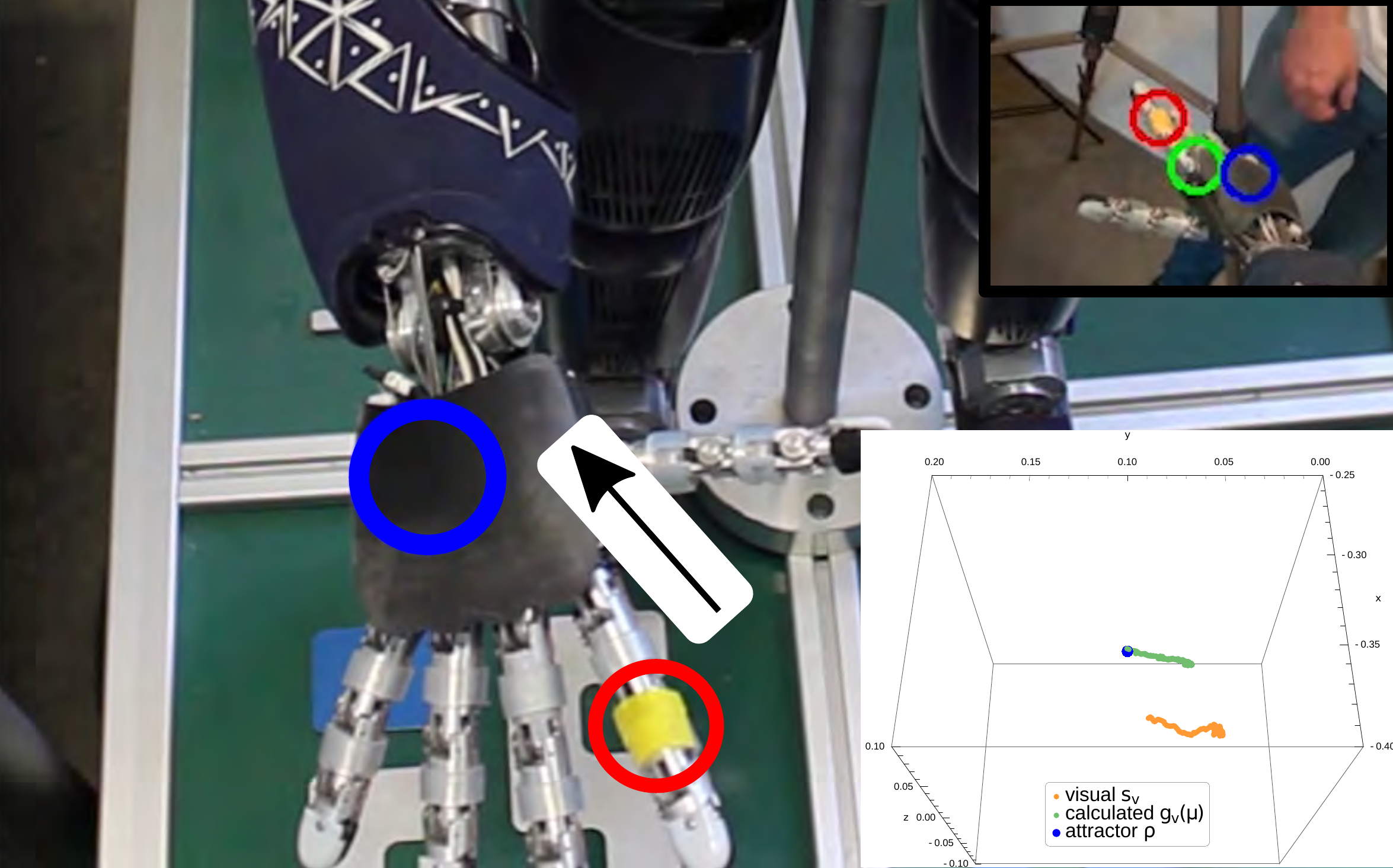}
		\label{results:adapt:finger}
	}	
	\subfigure[Visual marker on forearm.]{
		\centering
		\includegraphics[width=0.3\textwidth]{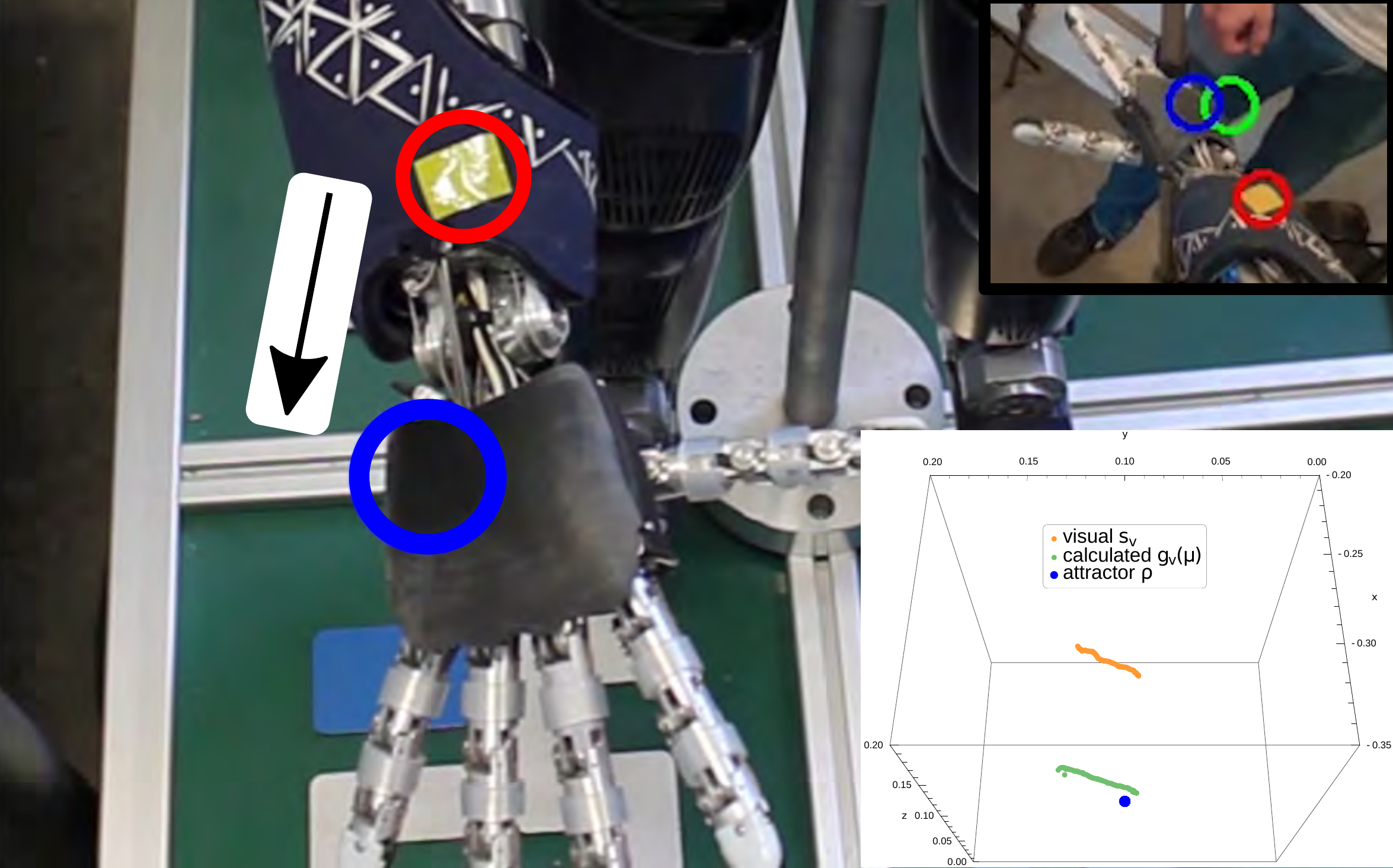}
		\label{results:adapt:forearm}
	}	
\end{minipage}
\hspace{-0.3cm}
\begin{minipage}{\textwidth}
        \subfigure[Jupiter experiment.]{
		\includegraphics[width=0.3\textwidth]{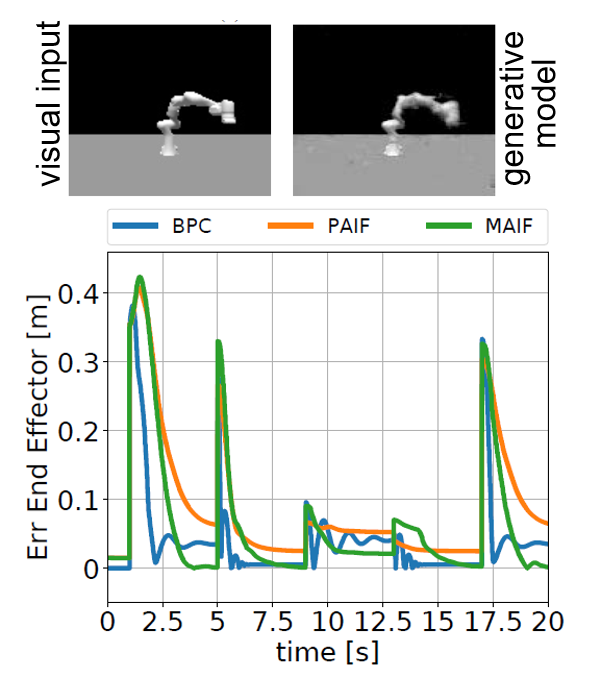}
		\label{results:adapt:jupiter}
		}
\end{minipage}
\caption{Adaptation: involuntary actions to fit changes in the world. (Left) In this experiment, we modified online the visual location of the arm end-effector (yellow marker) forcing the robot to adapt to the induced model-world mismatch. Taken from \citep{oliver2021empirical}. The algorithm produces counter-acting behaviours to maximize model evidence. Thus, it moves this new end-effector to the desired location. On the top right of each image, the left eye camera is shown. Desired end-effector position is shown in blue, visual perception in red and estimated position in green. The resulting direction of motion obtained from the algorithm is shown as an arrow. (Right) Jupiter experiment, taken from \citep{meo2021multimodal}. AIF approaches successfully adapt to the change in gravity when comparing with the built-in Panda controller (BPC). Multimodal AIF (MAIF) shows the best performance.}
	\label{results:adapt}
\end{figure}

This neuroscience-inspired approach to control the robot body brings adaptation in two ways: it counteracts external forces and changes in the body/environment without needing to relearn and it handles the trade-off between prior knowledge (acquired through experience) and pure reactive control driven by the sensory input. This is a great advantage of the AIF model with respect to classical control methods used in factories. The generalization to slightly different contexts is essential. In \cite{oliver2021empirical}, depicted in Fig. \ref{results:adapt}, we showed how when changing the visual end-effector location, the algorithm generated movements to adjust the end-effector to the robot internal model. This occurs because the algorithm computes the control actions by minimizing the variational free energy. Thus, it tries to reduce the difference between the internal model prediction and the end-effector visual observation. Recent experiments have shown evidence of similar behaviour in humans \citep{lanillos2020predictive}. This inherent mechanism is useful to deal with environmental and body changes, such as robot compliance or gravity changes. In \citep{meo2021multimodal} we described the Jupiter experiment on the Franka Emika Panda robot, which illustrates the advantages of an AIF model and the limitations of classical control methods used in factories. Figure \ref{results:adapt:jupiter} shows the performance comparison between the Panda built-in controller and two AIF controllers. Our controller presented equivalent accuracy when increasing the gravity to $g = 24.79  {m}/{s^2}$. Furthermore, AIF approaches have shown improved performance when comparing to state-of-the-art model predictive control and interesting properties for fault-tolerant control~\citep{pezzato2020active,pezzato2020novel}.


\section{Is this me? self-awareness}
\label{sec:awarenes}
Going from estimation and control to being aware of our body is a big leap. A breakthrough experiment, coined as the rubber-hand illusion (RHI, \cite{botvinick1998rubber}), showed how flexible humans are when perceiving their bodies. Participants felt a plastic arm as their arm in less than one minute by visuotactile stimulation. Several fMRI studies have looked into the neural correlates for these kinds of body-ownership illusions~\citep{makin2008other,hinz2018drifting}. Three areas were consistently found activated during the RHI: posterior parietal cortex (including the intra-parietal cortex and temporo-parietal junction), premotor cortex and lateral cerebellum. The cerebellum is assumed to compute the temporal relationship between visual and tactile signals, thus playing a role in the integration of visual, tactile and proprioceptive body-related signals. The premotor and intra-parietal cortex are multisensory areas, also integrating visual, tactile and proprioceptive signals present during the rubber hand illusion. Hence, the right crossmodal neural activations during sensorimotor integration may be the key for body-ownership. Within the predictive processing approach, we can answer a basic notion of `Is this me?' by evaluating whether the sensations fit the internal model~\citep{kahl2018predictive,lanillos2020robot}. 

\begin{figure}[hbtp!]
	\centering
    \includegraphics[width=1.0\linewidth]{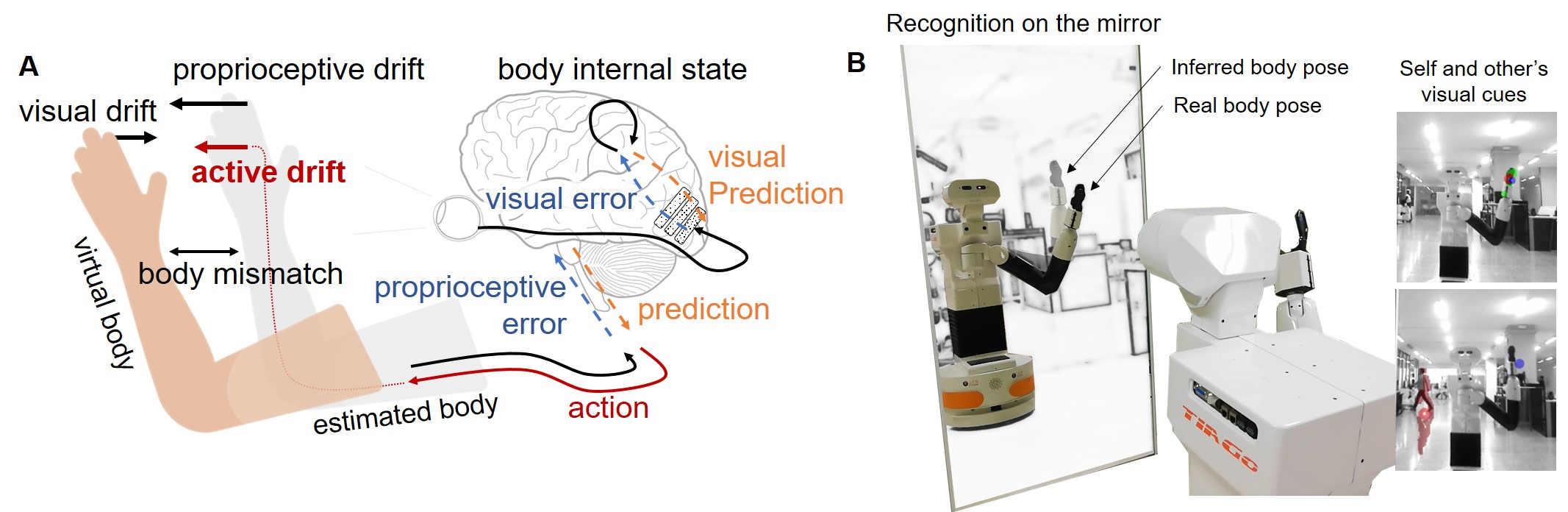}
	\caption{Towards human-like perception. (A) Human body perception and action when exposed to a virtual environment where the virtual body does not correspond to the actual body pose. (B) Modelling mirror self-recognition on a robot.}
	\label{fig:self}
\end{figure}

Cognitive robotics has also tried to enlighten the mechanisms behind body awareness~\citep{lanillos2017enactive,tani2020cognitive}. What we can already replicate are perceptual effects observed during the RHI \citep{hinz2018drifting, rood2020deep} due to the recalibration of the body posture when merging sensory conflicting information. Precisely, the real hand is mislocalized towards the plastic hand due to the illusion. Furthermore, we also showed evidence of an active component derived from the AIF hypothesis~\citep{lanillos2020predictive}. Figure \ref{fig:self}A describes the perceptual and active effects from the predictive brain point of view when exposing a person to a virtual body.

Furthermore, in~\citep{lanillos2020robot}, Fig.~\ref{fig:self}B, we showed how to use AIF to solve a simple form of artificial self-recognition~\citep{hoffmann2021robot} by visual-kinesthetic matching using movement cues. In essence, the robot infers that it is itself by evaluating again the model evidence through the VFE. $\mathcal{F}$ is therefore used as a bound on how much the observations fit the internal model. In other words, how much the sensations are being produced by its body.

Finally, `Did I do it?'~\citep{haggard2017sense} is still an evasive question. Although it is already possible to empirically evaluate the experience of agency and some theoretical models have been proposed~\citep{blakemore2000can,hommel2015action}, yet no computational model can properly replicate the process. Unveiling the mechanism of being aware of the effects that we generate in the world is critical for robotics as it provides safe interaction in complex environments and very relevant for imitation and social interaction \citep{wirkuttis2021controlling}.

\section{Conclusion}
\label{sec:conlusions}

We discussed some relevant models and experiments in robotics based on neuroscientific theories of how humans perceive their bodies. In particular, we have presented a low-level estimation and adaptive control algorithm that takes inspiration from how the brain may process sensory information and generates actions through the minimization of surprise. Hence, we make a connection between predictive processing theory in cognitive (neuro)science and outstanding challenges in the field of robotics. Body intelligence is one of the biggest challenges described by Moravec's paradox that is not yet solved~\citep{moravec1988mind}. The described proof-of-concept experiments show how understanding the brain can revolutionize robotics and embodied artificial intelligence in terms of adaptation, generalization, flexibility and robustness. As such, this research marries ideas in neuroscience and artificial intelligence with the aim of developing a new generation of naturally intelligent systems~\citep{VanGerven2017b}.


\bibliography{selfception}
\bibliographystyle{iclr2021_conference}


\end{document}

%% file: math_commands.tex
\usepackage{amsmath,amsfonts,bm}

\def\eqref#1{equation~\ref{#1}}

\def\1{\bm{1}}

\DeclareMathAlphabet{\mathsfit}{\encodingdefault}{\sfdefault}{m}{sl}
\SetMathAlphabet{\mathsfit}{bold}{\encodingdefault}{\sfdefault}{bx}{n}

